\pdfoutput=1
\documentclass[11pt]{article}

\usepackage[review]{acl}
\usepackage{amssymb}
\usepackage{times}
\usepackage{multirow}
\usepackage{latexsym}
\usepackage{listings}
\usepackage{wrapfig}
\usepackage{tcolorbox}
\tcbuselibrary{skins}
\usepackage[T1]{fontenc}

\usepackage{caption}
\usepackage[utf8]{inputenc}

\usepackage{microtype}

\usepackage{inconsolata}

\usepackage{graphicx}
\usepackage{amsmath}

\usepackage{algorithm}
\usepackage{algpseudocode}
\usepackage{tikz}
\usepackage{multicol}
\usepackage{subcaption}
\usepackage{amsfonts}

\usepackage{tabularx, booktabs}
\newcolumntype{Y}{>{\centering\arraybackslash}X}

\usetikzlibrary{fit,calc}
\usepackage{xcolor}

\colorlet{myshadow}{gray!40}

\usepackage{enumitem}
\title{Gradient-Aware Weight Quantization for Large Language Models}

\author{
  \textbf{Yihua Shao\textsuperscript{1,2}},
  \textbf{Yan Gu\textsuperscript{3}},
  \textbf{Siyu Chen\textsuperscript{4}},
  \textbf{Haiyang Liu\textsuperscript{3}}, 
  \textbf{Zixian Zhu\textsuperscript{4}}, 
  \textbf{Zijian Ling\textsuperscript{4}},
  \textbf{Minxi Yan\textsuperscript{2}},\\
  \textbf{Ziyang Yan\textsuperscript{5}},
  \textbf{Chenyu Zhang\textsuperscript{5}},
  \textbf{Michele Magno\textsuperscript{6}},
  \textbf{Haotong Qin\textsuperscript{6*}},\\
  \textbf{Yan Wang\textsuperscript{3}},
  \textbf{Jingcai Guo\textsuperscript{7}},
    \textbf{Ling Shao\textsuperscript{1*}}
  \textbf{Hao Tang\textsuperscript{1*}}
\\
\\
  \textsuperscript{1}PKU \quad
  \textsuperscript{2}CASIA \quad
  \textsuperscript{3}THU \quad
  \textsuperscript{4}USTB \quad
  \textsuperscript{5}UNITN  \quad
  \textsuperscript{6}ETHz \quad
  \textsuperscript{7}PolyU \quad
  \textsuperscript{8}UCAS \quad
\\
\thanks{The first two authors contributed equally. }
    \textbf{correspondence authors:} \href{mailto:email@domain}{qinhaotong@gmail.com, haotang@pku.edu.cn}
}

\begin{document}
\maketitle
\begin{abstract}
Large language models (LLMs) show impressive performance in solving complex language tasks. However, its large number of parameters presents significant challenges for the deployment. So, compressing LLMs to low bits can enable to deploy on resource-constrained devices. To address this problem, we propose gradient-aware weight quantization (GWQ), the first quantization approach for low-bit weight quantization that leverages gradients to localize outliers, requiring only a minimal amount of calibration data for outlier detection. GWQ retains the top 1\% outliers preferentially at FP16 precision, while the remaining non-outlier weights are stored in a low-bit. We widely evaluate GWQ on different task include language modeling, grounding detection, massive multitask language understanding and vision-language question and answering. Results show that models quantified by GWQ performs better than other quantization method. During quantization process, GWQ only need one calibration set to realize effective quant. Also, GWQ achieves 1.2$\times$ inference speedup in comparison to the original model and effectively reduces the inference memory.
\end{abstract}

\section{Introduction}

Large language models (LLMs) \citep{touvron2023llama,achiam2023gpt,almazrouei2023falcon,touvron2023llama2openfoundation} based on Transformer \citep{vaswani2017attention} have demonstrated their outstanding capabilities in scenarios such as language modeling \citep{brown2020language}. Its ability to handle complex linguistic tasks is due to its large pre-trained data \citep{kaplan2020scaling, yan20243dsceneeditor,yan2025learning} and an astronomical number of parameters \citep{chowdhery2023palm}. However, the huge memory consumption leads to great difficulties in its application and deployment \citep{li2024personal, remondino2023critical}. Therefore, when deploying large language models, there are often huge GPU clusters to support the inference of the models \citep{xu2024survey}.

In order to apply LLMs to resource-constrained edge devices \citep{li2024quasar,huang2024slim}, model compression \citep{ma2023llm,gu2023knowledge} has become common means to reduce the computational memory. Among them, post-training quantinazation of LLMs \citep{dettmers2023case} is a widely used method. Most current quantization methods target to compress modela to 3 or 4 bits \citep{dettmers2023spqr,frantar2022gptq}, while ensuring models has less loss of performance. Therefore, reducing the size of the model while ensuring its performance is necessary by design effective quantization algorithms \citep{lin2024awq}. With effective quantization algorithms, researchers can make low-bit models' performance close to that of 16-bit models.

OBQ \citep{frantar2022optimal} indicates that when the pre-trained model has converged, the model should exhibit zero gradients. However, our experiments reveal that the LLMs still generate gradients in response to different text inputs. Inspired by this phenomenon, we propose a gradient-aware post-training weight-only quantization method called GWQ. GWQ is the first post-training quantization approach to utilize gradients to locate outliers in pre-trained models. As same as other quantization method~\cite{lin2023awq,dettmers2023spqr}, GWQ selects 1\% sensitive weights as outliers by searching the gradient for the response to the calibration set, preserving these outliers in FP16 precision while quanting the remaining 99\% of the weights to 4 bits or 3 bits.

In summary, the primary contributions of this paper are as follows:
\begin{itemize}[noitemsep, topsep=0pt]
\item [1)] 
GWQ discovered that locating sensitive weights by first-order gradient is more rational than Hessian matrix.
\item [2)] 
GWQ is the first accurate first-order gradient-aware post-training weight quantization method for pre-trained LLMs, requiring only one of calibration data to locating outliers.
\item [3)] 
GWQ achieves state-of-the-art (SOTA) in language modeling, grounding detection, massive multitask language understanding and vision-language question and answering tasks in different LLMs. Meanwhile, the quantified models have achieved 1.2$\times$  acceleration compared to the original model and cost less memory during inference.
\end{itemize}

\section{Related Work}

\noindent\textbf{Post-training Quantization.}
Post-training quantization (PTQ) applies mainly to visual models \citep{gholami2022survey}. For example, AdaRound \citep{nagel2020up}, AdaQuant \citep{hubara2020improving}, and BitSplit \citep{wang2024data} can be used for models with less parameters. When models have a large number of parameters, they cause a significant loss but the reduction in memory is not significant. Most quantization methods are developed by Optimal Brain Quantization (OBQ) \citep{frantar2022optimal}. OBQ is developed by the Optimal Brain Surgeon (OBS) \citep{hassibi1993optimal,singh2020woodfisher,frantar2021m}, which default a model response to input is 0 at the end of training. But the OBQ applies weight pruning to model quantization by quantifying the weights that have the less impact on the network. Some PTQ methods locating outlier by reinforcement learning \citep{wang2024data} to retrieve the best outlier on a calibration set with a very large sample size. However, the reinforcement learning is not robust and is difficult to converge compared to other PTQ methods, making the calibration process inefficient.

\noindent\textbf{Large Language Model Quantization.}
Currently, large language model quantization is mainly divided into quantization-based training \citep{liu2023llm,bondarenko2024low,zhu2023pad, shao2025context, shao2025tr} and post-training quantization. Quantization during training is very difficult to apply due to the large amount of computational resources and overhead required, as well as the large amount of training data. The remaining methods such as ZeroQuant \citep{yao2022zeroquant} utilize knowledge distillation to train one transformer layer at a time instead of training all transformer layers directly. However, poor quality of certain data can also affect the performance of the model after quantization \citep{wang2024generated}. Post-training quantization requires less data compared to quantization-aware training, and only a few thousand or even a few hundred calibration sets are needed to complete the retrieval of sensitive weights for the model. This greatly reduces the data cost and the computational cost of model quantization. Both GPTQ \citep{frantar2022gptq} and SPQR \citep{dettmers2023spqr}, quantization methods derived from the OBQ method, have shown superior performance in large language models. There are also methods such as AWQ \citep{lin2024awq}, LLM.int8() \citep{dettmers2022gpt3} that determine the outlier of the model by searching for the activation of the model, and then quantify it with the corresponding method. However, the method of searching for outliers by activation has poor interpretability.

\begin{figure*}[htbp]    % 常规操作\begin{figure}开头说明插入图片
% 后面跟着的[htbp]是图片在文档中放置的位置，也称为浮动体的位置，关于这个我们后面的文章会聊聊，现在不管，照写就是了
  \vspace{-0.8cm}
  \centering            % 前面说过，图片放置在中间
  \subfloat[Outliers of Layer 18 SPQR]   % 第一张子图的下标（注意：注释要写在[]中括号内）
  {
      \label{fig:subfig1}\includegraphics[width=0.24\textwidth]{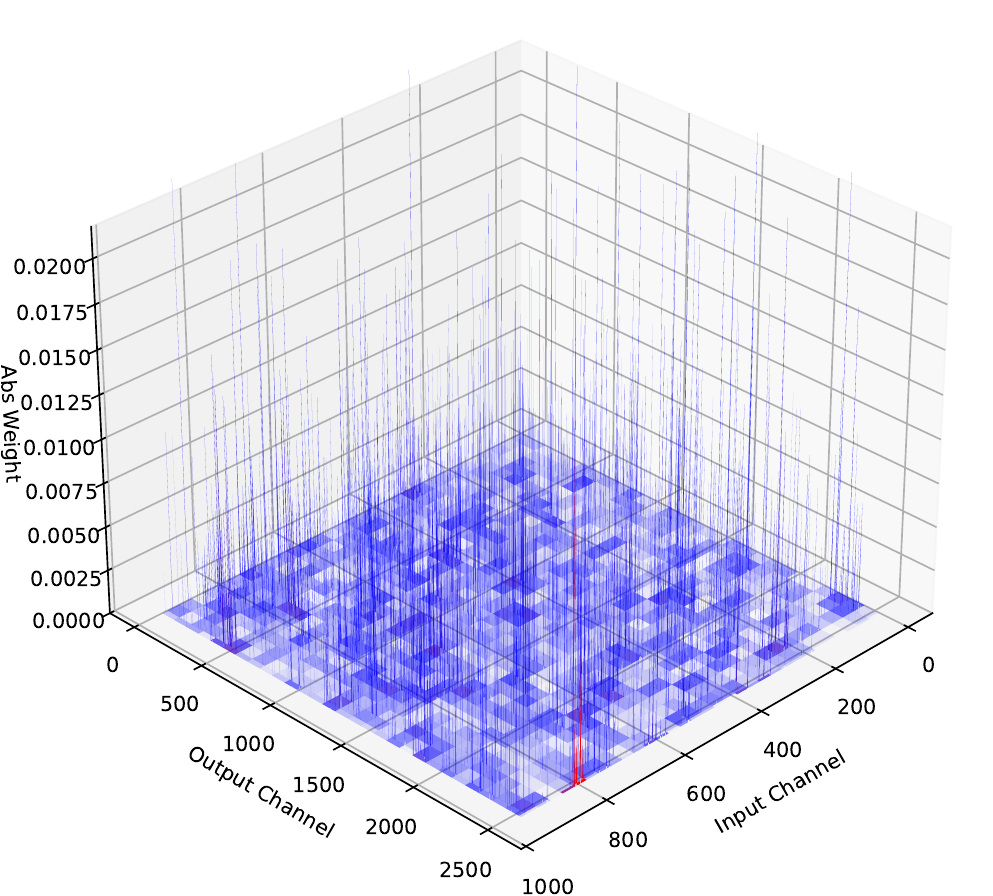}
      
  }
  \subfloat[Outliers of Layer 18 GWQ]
  {
      \label{fig:subfig2}\includegraphics[width=0.24\textwidth]{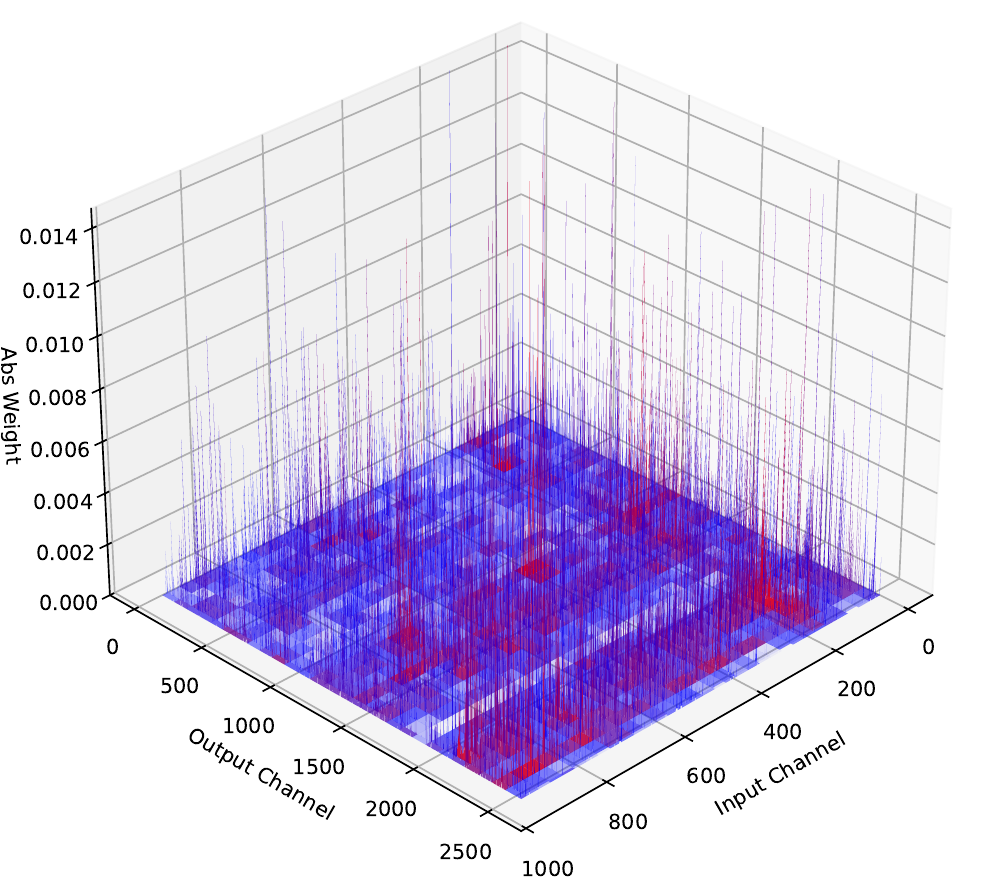}
  }
  \subfloat[Outliers of Layer 30 SPQR]
  {
      \label{fig:subfig3}\includegraphics[width=0.24\textwidth]{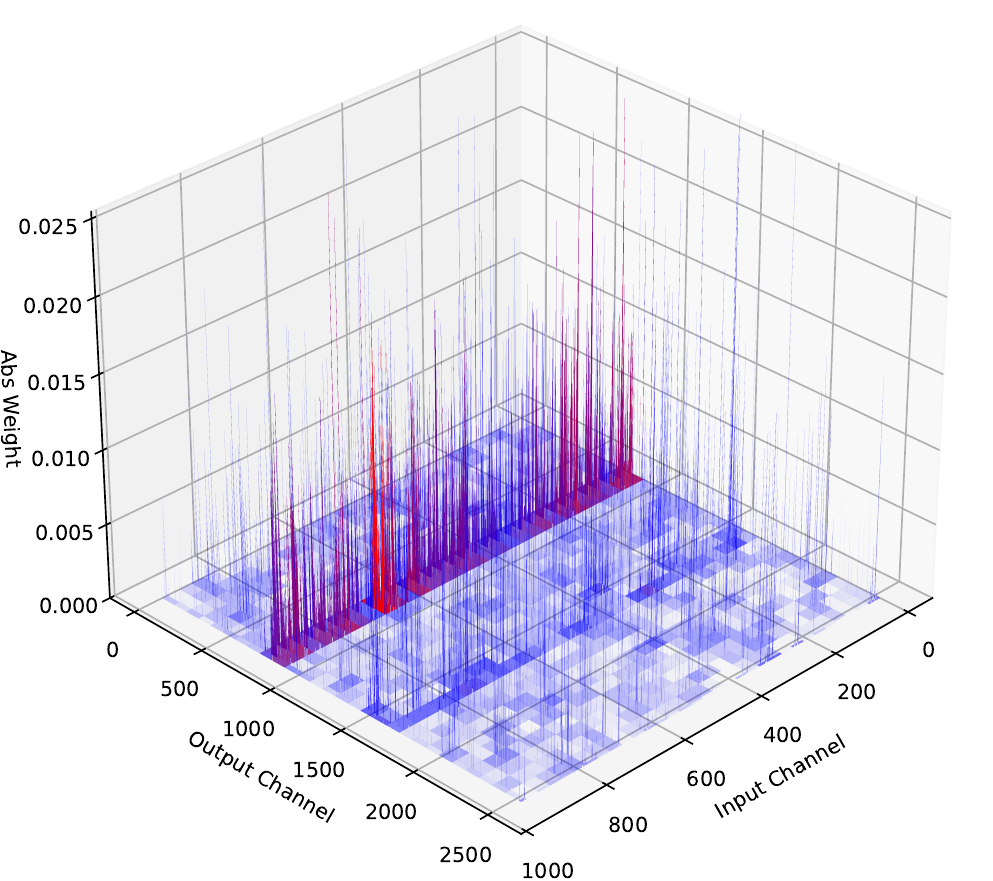}
  }
  \subfloat[Outliers of Layer 30 GWQ]
  {
      \label{fig:subfig4}\includegraphics[width=0.24\textwidth]{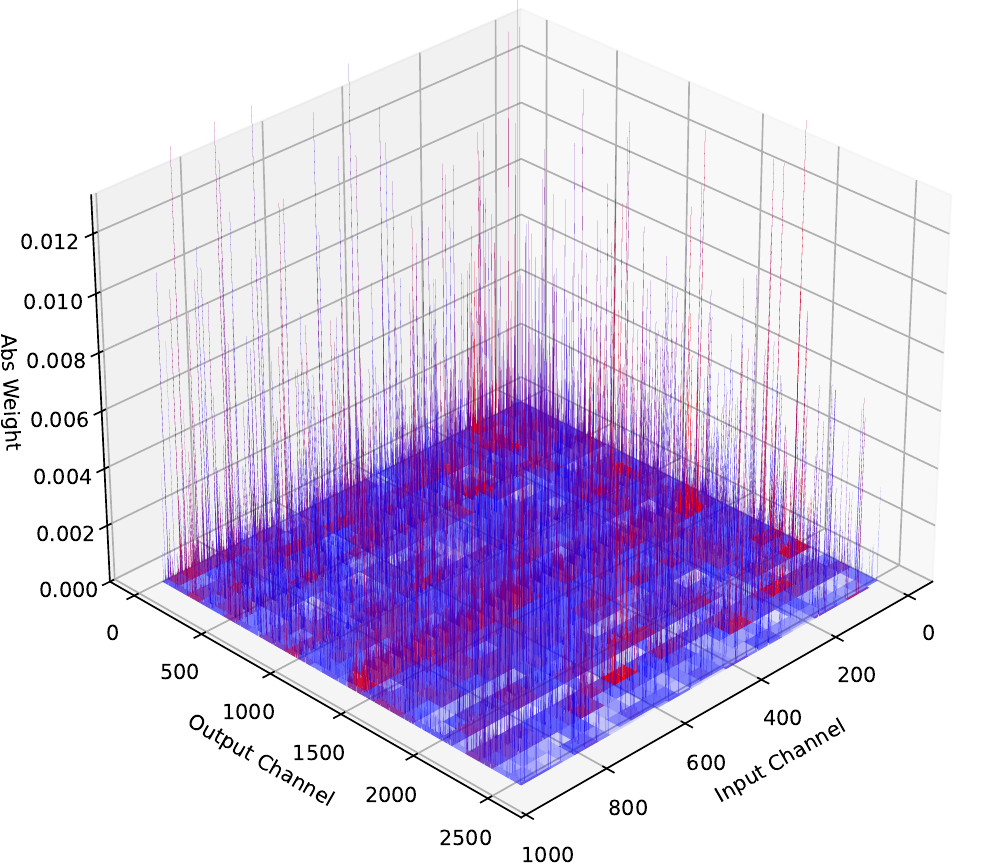}
  }
  \caption{\textbf{Compared of GWQ and SPQR.} Compared to SPQR, the outliers searched by GWQ are much sparser and easier to quantization.}    % 整个图片的说明，注释写在{}内
  \label{fig:subfig_1}            % 整个图片的标签编号，注意这里跟子图是一样的道理，标签不能重复 
  \vspace{-0.4cm}
\end{figure*}
\begin{table*}[htbp]
\caption{\textbf{Gradient quantization of non-outliers with Hessian in comparison on C4 dataset \citep{raffel2020exploring}.} We localize outliers by gradient and Hessian and quant the non-outliers with RTN \citep{nagel2020up} to 4 bits.}
\label{Table0}
 \centering
\resizebox{0.9\linewidth}{!}{
\begin{tabular}{c|c|ccc|ccc}
\toprule 
& & \multicolumn{3}{c|}{\textbf{FP16 (\%) base on Gradients}}  
 & \multicolumn{3}{c}{\textbf{FP16 (\%) base on Hessian}}\\
% \cmidrule(lr){2-4} \cmidrule(l){5-7}
        \textbf{PPL$\downarrow$}& \textbf{FP16}  & 0.1\%  & 0.5\%  & 1\%  & 0.1\% & 0.5\% & 1\%  \\  \midrule
        
        Llama-2-7b-hf & 6.9759 & 20.5896 & 16.8329 & 15.4319 & 24.0418 & 20.2053 & 18.5902  \\% \hline
        Llama-2-13b-hf & 6.4734 &20.3549 & 16.6689 & 15.1590 & 23.9322 & 20.1040 & 18.3405  \\
        Llama-3-8b-hf & 8.9990 &22.4648 & 18.9843 &17.5901 & 22.4423 & 22.9650 & 19.3283  \\
\bottomrule
\end{tabular}}
\vspace{-0.4cm}
\end{table*}
\section{The Proposed Method}
\label{methods}

% \begin{table*}[ht]
% \caption{\textbf{Perplexity of Llama-2-hf family and Llama-3-8B-hf on WikiText datasets \citep{merity2016pointer}.} GWQ outperforms other quantization methods, especially when the average bit-width of the model is significantly lower.}
% \label{Table1}
% \begin{center}
%     \centering
%     % \begin{tabular}{c|c|c|c|c|c|c|c}
%     % \resizebox{220pt}{!}{
    
%     \begin{tabular}{cccccccc}
%     \hline
%         % \multirow{2}{*}{PPL$\downarrow$}& \multirow{2}{*}{Avg\_bit}  & 
%         & &
%         \multicolumn{3}{c}{\textbf{FP16(\%) base on Gradient}}  & \multicolumn{3}{c}{\textbf{FP16(\%) base on Hessian}}\\ 
%         % \cline{3-8}
%         \textbf{PPL$\downarrow$}& FP16  & 0.1\%  & 0.5\%  & 1\%  & 0.1\% & 0.5\% & 1\%  \\ \hline
        
%         Llama-2-7b-hf & - & - & 5.47 & 6.97 & 4.88 & 6.47 & 6.23  \\% \hline
%         Llama-2-13b-hf & - &- & 5.83 & 7.79 & 5.12 & 7.72 & 8.21  \\
%         Llama-3-8b-hf & - &- & 5.83 & 7.79 & 5.12 & 7.72 & 8.21  \\
%         %\hline

%          \hline
%     \end{tabular}
%     % }
% \end{center}
% \end{table*}

The outlier search method based on Hessian matrix \citep{frantar2022gptq,dettmers2023spqr} is built upon the OBQ approach \citep{frantar2022optq}. The OBQ method retrieves the parameters that have the least impact on the model parameters could be expressed as Eq. \eqref{E}:
% \begin{small}
\begin{equation}
\label{E}
    \begin{split}
        \Delta E=\sum_{i} g_{i} \Delta w_{i}+\frac{1}{2} \sum_{i} h_{i i} \Delta w_{i}^{2}+ 
        \\
        \frac{1}{2}\sum_{i \neq j} h_{i j} \Delta w_{i} \Delta w_{j}+O\left(\Delta w^{3}\right),
    \end{split}
\end{equation}

\noindent where $g_{i} =\frac{\partial E }{\partial w_{i} }$ is the first order gradient. 
In the OBQ framework, it is suggested that once a pre-trained model has fully converged, its gradients $g_{i} =\frac{\partial E }{\partial w_{i} }$  should ideally approach zero. 

However, our experiments demonstrate that LLMs continue to generate gradients in response to various text inputs.
\citet{bondarenko2023quantizable} denotes that the occurrence of outliers in large models arises from attention heads attempting to avoid updating hidden states. During this process, the softmax function magnifies the formation of strong outliers. Building on this observation, we hypothesize that when a well-trained LLM computes gradients for text input, it often focuses these gradients on irrelevant outliers as a mechanism to prevent hidden state updates. As shown in Fig. \ref{fig:subfig_1}, compared to SPQR, GWQ demonstrates a sparser allocation of outliers. Specifically, the outliers identified are distributed either per channel or per row, which contrasts with the more dense and uniform distribution of outliers observed in SPQR. This sparsity in GWQ results in a more reasonable allocation, as it does not require concentrating on the compensation of sensitive weight errors across every layer output. As shown in Tab. \ref{Table0}, with different proportions of outliers, the model using Hessian matrix search and quantized using RTN \citep{nagel2020up} performs worse than the model using gradient search and quantized using RTN on the C4 dataset \citep{raffel2020exploring}, so it can be learned that it is more reasonable to use the gradient to do outliers localization.

\begin{figure*}
\vspace{-0.6cm}
    \centering
    \includegraphics[width=1.0\linewidth]{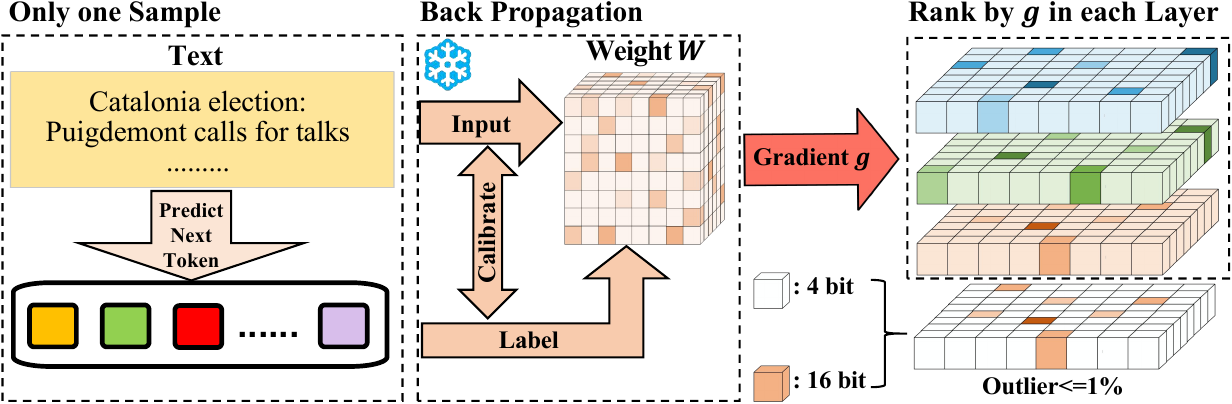}
    \caption{\textbf{Sensitive weight Location. }GWQ utilizes a single calibration sample to calibrate and backpropagates this sample with the subsequent token from the large model's output as the target label. By halting the backpropagation process, GWQ is able to capture the model gradient $g$ prior to the update of the weights $W$. Within each layer, the gradients associated with each weight block are sorted in relation to their respective weights, and the top 1\% with the largest $\left | g \right | $ are identified as the model's outliers, which are the weights that are particularly sensitive.}
    \label{fig:pipeline}
    \vspace{-1.0em}
\end{figure*}

\subsection{Sensitive Weight Location}
GWQ captures the absolute gradients with respect to LLM, acquired through back-propagation, the detail is shown in Alg. \ref{gradient_alg}. We disabled bias in all selected linear layers since the bias term does not get multiplied by inputs in the same way as weights do. This can cause the weight gradients to be unrepresentative of the actual importance of the weights, leading to a disproportionate influence on the MSE loss calculation. As it shown in Fig. \ref{fig:pipeline}, the calibration dataset is represented as $D_c$, the weights of LLM as $W$, and the loss function of LLM as $\mathcal{L} = (W; D_c)$. So, the gradients $g$ of the LLM could be expressed as \eqref{grad_obtaining},
\begin{equation}
    g = \bigtriangledown_W \mathcal{L} = \bigtriangledown_W \mathcal{L} (W; D_c),\label{grad_obtaining}
\end{equation}

In order to locate the sensitivities as outliers, we calculate loss $\mathcal{L}(W_Q)$ after model quantization based on Eq. \eqref{Taylor}.
\begin{equation}
    \mathcal{L}(W_Q) \simeq \mathcal{L}(W) - g^\top (W-W_Q). \label{Taylor}
\end{equation}
We define the quantization process as an optimization problem to search for quantitatively optimal outliers. The process can be expressed as Eq. \eqref{Q},
\begin{equation}
    Q\left (W  \right ) =arg\min_{Q} \left \| W-W_{Q}  \right \| _{2}^{2}, \label{Q}
\end{equation}
Therefore the final weight optimization formula can be expressed as Eq \eqref{final_OPO},
\begin{equation}
   Q\left (W  \right ) =arg\min_{Q}\bigtriangledown_W \mathcal{L} (W; D_c)(W-W_{Q}). \label{final_OPO}
\end{equation}

\subsection{Gradient Aware Quantization}

To quantize the LLM weight $W$, we first compute the scale $s$ and the zero point $z$, which are calculated as depicted in Eq. \eqref{s} and \eqref{z}, where $\beta$ represents the group size (set to 16 in this case) and $b$ is the bit-width for quantization.
%\begin{equation}
%    \begin{aligned}
%    s = \frac{\max(W) - \min(W)}{2^{b-1}},
%    \end{aligned} \label{s}
%\end{equation}
%\begin{equation}
%\begin{aligned}
%    z = \frac{-\min{W}}{s}.
%    \end{aligned}\label{z}
%\end{equation}
\begin{equation}
    s = \frac{\max(W) - \min(W)}{2^{b-1}}, \label{s}
\end{equation}
\begin{equation}
        z = \frac{-\min{W}}{s}. \label{z} 
\end{equation}
We select the weight by absolute values of the weight gradients $|W_{grad}|$, the largest 1\% of them are filtered as outliers, and the rest are used as quantization weights. The process can be expressed as Eq. \eqref{o},
\begin{equation}
        O = \mathbb{I} \left( |W_{grad}| \geq \text{quantile}(\alpha_1) \right),
\label{o}
\end{equation}
where $\alpha_1$ is the specific quantile value of the magnitude of weight gradients, which are used to identify outliers matrix $O$. 

The quantization and dequantization operations are carried out per channel within each group to assess the quantization error. The formulas for quantization and dequantization are provided below (Eq. \eqref{q} and \eqref{w}), and any identified outliers are excluded from the error computation Eq. \eqref{o}:
\begin{equation}
            q = round(\frac{W}{s}) + z,
\label{q}
\end{equation}

Finally, the quantized weight $Q(W)$ can be denoted as Eq \eqref{w}:
\begin{equation}
            Q(W) = s \times (q-z).\label{w} 
\end{equation}

\section{Experiments}
\label{experiment}

\begin{table*}[]
\vspace{-0.4in}
\caption{\textbf{Results of Llama family on language modeling task.} GWQ outperforms other quantization methods, especially when the average bit-width of the model is significantly lower.}
\label{Table1}
    \centering
\resizebox{\linewidth}{!}{
\begin{tabular}{c|cc|cccccc}
\toprule 
         & \textbf{Average}   & \textbf{Calibration} & \multicolumn{2}{c}{\textbf{Llama-2-7B-hf}} & \multicolumn{2}{c}{\textbf{Llama-2-13B-hf}} & \multicolumn{2}{c}{\textbf{Llama-3-8B-hf}} \\
\textbf{PPL$\downarrow$ / Acc$\uparrow$} & \textbf{bit}& \textbf{number}      & WikiText2           & C4          & WikiText2           & C4           & WikiText2           & C4          \\     \midrule
   Original & 16 & - & 5.47 / 61.01 & 6.97 / 56.77 & 4.88 / 62.88 & 6.47 / 57.90 & 6.23 / 58.34 & 8.99 / 51.71 \\\midrule
+ OPTQ & 4 &1024 & 9.76 / 51.53 & 	11.24 / 49.01 & 9.37/ 51.89 & 10.65 / 50.02 & 11.56 / 39.34 &15.84 /44.93\\ %\hline
   
 + QuIP & 4 &1024 & 8.43 / 53.13 & 	10.01 / 50.82 &8.13 / 53.48& 9.49 / 51.33 & 10.89 / 40.84 & 14.03 / 46.22\\ %\hline
+ GPTQ & 4 &1024 & 5.83 / 59.88 & 7.79 / 54.74 & 5.12 / 61.91 & 7.72 / 56.18 & 8.21 / 43.78 & 11.48 / 47.75\\ %\hline
        
        + AWQ & 4 & 512 & 5.60 / 60.54 & 7.70 / 55.16 & 4.97 / 62.53 & 7.10 / 56.44  & 6.64 / 57.07 &  10.69 / 47.75 \\ \midrule
        + SPQR-O & 4.63 & 1024 & 5.53 / 60.80 & 7.02 / 56.63 & 4.93 / 62.71  & 6.50 / 57.78 & 6.42 / 57.70 & 9.23 / 51.20\\ %\hline
        + SPQR-R & 3.98 & 1024 & 5.55 / 60.56 & 7.06 / 55.18 & 4.95 / 62.55  & 6.53 / 56.45 & 6.45 / 57.09 & 9.26 / 48.97\\ %\hline
        + GWQ-O (Ours) & 4.63 & \textbf{1} & \textbf{5.48 / 60.87} & \textbf{6.99 / 56.72} & \textbf{4.90 / 62.76} & \textbf{6.48 / 53.84} &\textbf{6.40 / 57.83} & \textbf{9.11 / 51.46} \\ %\hline
        + GWQ-R (Ours) &3.98 & \textbf{1} & \textbf{5.50 / 60.59} & \textbf{7.01 / 56.20} & \textbf{4.38 / 62.56} & \textbf{6.50 / 56.77} & \textbf{6.41 / 57.69} & \textbf{9.15 / 49.89}\\   \bottomrule                  
\end{tabular}}
\vspace{-0.4cm}
\end{table*}

\begin{table*}[]
%\vspace{-0.4in}
\caption{\textbf{Results of QWEN-VL families on language molding task and grounding detection task.} GWQ outperforms other quantization methods on both language modeling task and grounding detection task.}
\label{Table2}
    \centering
\resizebox{\linewidth}{!}{
\begin{tabular}{c|cc|ccccccccc}
\toprule 
         & \textbf{Average} & \textbf{Calibration} & \multicolumn{4}{c}{\textbf{QWEN-VL}}                            & \multicolumn{5}{c}{\textbf{QWEN-VL-chat}}                                             \\
\textbf{PPL$\downarrow$ / Acc$\uparrow$} & \textbf{bit}   & \textbf{number}      & \multicolumn{2}{c}{WikiText2}  & C4          & RefCOCO & \multicolumn{2}{c}{WikiText2}  & C4            & \multicolumn{2}{c}{RefCOCO} \\ \midrule
Original & 16&-& \multicolumn{2}{c}{8.21 / 54.18} & 10.10 / 49.51 & 89.36   & \multicolumn{2}{c}{9.85 / 52.06} & 11.87 / 48.07   & \multicolumn{2}{c}{88.55}   \\ \midrule

+ OPTQ & 4 &1024          & \multicolumn{2}{c}{9.34 / 53.13} & 12.99 /45.76 & 85.78   & \multicolumn{2}{c}{13.78 / 50.02} & 15.71 / 44.93   & \multicolumn{2}{c}{84.12}   \\

+ QuIP & 4 &1024          & \multicolumn{2}{c}{9.48 / 53.87} & 12.68 / 46.41 & 86.92   & \multicolumn{2}{c}{12.33 / 51.65} & 13.98 / 45.76   & \multicolumn{2}{c}{85.41}   \\

+ GPTQ & 4 &1024          & \multicolumn{2}{c}{9.31 / 52.07} & 11.97 / 47.08 & 87.13   & \multicolumn{2}{c}{11.50 / 52.06} & 13.59 / 46.09   & \multicolumn{2}{c}{86.67}   \\

+ AWQ & 4 &512          & \multicolumn{2}{c}{8.69 / 53.58} & 10.98 / 48.12  & 84.51    & \multicolumn{2}{c}{11.02 / 50.77} & 12.88 / 47.43   & \multicolumn{2}{c}{87.77}   \\ \midrule

+ SPQR-O& 4.63 &1024          & \multicolumn{2}{c}{8.54 / 54.05} & 10.25 / 49.26   & 88.47   & \multicolumn{2}{c}{10.13 / 51.99} & 12.02 / 47.35   & \multicolumn{2}{c}{88.26}   \\ 

+ SPQR-R &3.98 &1024          & \multicolumn{2}{c}{8.69 / 53.57} & 10.98 / 48.14  & 88.00   & \multicolumn{2}{c}{11.03 / 50.80} & 12.88 / 46.84   & \multicolumn{2}{c}{87.78}   \\ 

+ GWQ-O & 4.63 &\textbf{1}         & \multicolumn{2}{c}{\textbf{8.24 / 54.06}} & \textbf{10.10 / 49.50} & \textbf{88.98}    & \multicolumn{2}{c}{\textbf{9.86 / 52.03}} & \textbf{11.88 / 48.07}   & \multicolumn{2}{c}{\textbf{88.49}}   \\ 

+ GWQ-R & 3.98 &\textbf{1}          & \multicolumn{2}{c}{\textbf{8.53 / 53.87}} & \textbf{10.01 / 49.12}  & \textbf{88.42}   & \multicolumn{2}{c}{\textbf{10.34 / 51.93}} & \textbf{12.62 / 47.18}  & \multicolumn{2}{c}{\textbf{87.93}}   \\ \bottomrule
\end{tabular}}
\vspace{-0.4cm}
\end{table*}

\begin{figure}[!tbp]
    \centering
    \includegraphics[width=1.0\linewidth]{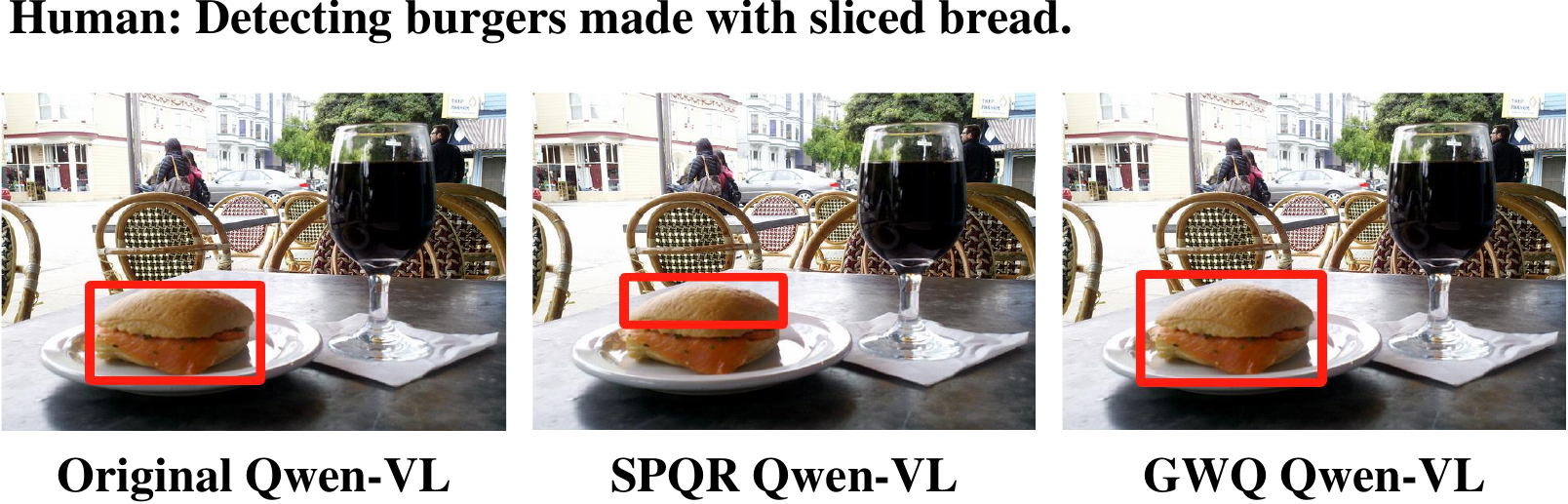}
    \caption{\textbf{Performance of Different Quantized Qwen-VL. }GWQ performs well compared to other quantization methods in the zero-shot grounding detection task.}
    \label{fig:QWEN}
\end{figure}

\textbf{Overview.} 
We first apply GWQ to the Llama2 \cite{touvron2023llama2openfoundation} and Llama3 \cite{dubey2024llama} family of models and evaluate its feasibility on a language modeling task. We then extend our approach to multimodal models \cite{bai2023qwen,bai2023qwenvlversatilevisionlanguagemodel,liu2024visual}, measuring the generalizability of our approach to multimodal scenarios with different base models for the grounding detection task, vision language question and answer (VQA). To ensure the fairness of the experiments, we compressed the average number of bits of all the mixed-precision quantized models to less than 4 bits.

\subsection{Experiments Setting}
\label{setting}
\textbf{Setups.} 
The language-only models we selected include Llama2-7B and 13B \citep{touvron2023llama2openfoundation} and Llama3-8b \cite{dubey2024llama}. As for vision-language model, we chose LLaVA-V1.5 \cite{liu2024visual}, Qwen-VL \cite{bai2023qwen,bai2023qwenvlversatilevisionlanguagemodel}, Qwen-VL-chat \cite{bai2023qwen,bai2023qwenvlversatilevisionlanguagemodel}.
These models were quantized using the first sample of the RedPajama dataset \cite{weber2024redpajama}. To evaluate Perplexity (PPL) and Accuracy (Acc.), we utilize the WikiText2 \citep{merity2016pointer} and C4 \citep{raffel2020exploring} as validation sets. 
All experiments were conducted on an NVIDIA A800 (80G) GPU.

\noindent \textbf{Evaluation Metrics.}
For the language modeling task, we focus on the model's perplexity (PPL$\downarrow$) and the next token prediction accuracy (Acc$\uparrow$). Model's perplexity can be expressed as Eq \eqref{eq:ppl},
\begin{equation}
    \label{eq:ppl}
    \text{PPL} = \exp\left(-\frac{1}{N} \sum_{i=1}^N \log P(x_i \mid x_{0: i-1})\right), 
\end{equation}
where $x_i$ represents the predicted token conditioned on the former tokens ($x_{<i}$) in inference step $i$ and $c_w$ denotes the average characters per word in the generated sequence.

The next token prediction accuracy can be expressed as Eq. \eqref{acc},
\begin{equation}
\label{acc}
    \mathrm{Acc} =\frac{N_{\hat{y_t}=\overrightarrow{y_{t} }  }  }{T} ,
\end{equation}
where $\hat{y_t}$ is the predicted token by LLMs, $\overrightarrow{y_{t}}$ is the ground truth token, $T$ is the number of all tokens.

The exact calculation of next token prediction accuracy will be explained in \ref{C_acc}.

\noindent \textbf{Baselines.} 
\label{baseline}
Our baseline is categorized into uniform precision quantization and mixed precision quantization. For uniform precision quantization, we choose OPTQ \cite{frantar2022optq}, QUIP \cite{chee2023quip}, GPTQ \cite{frantar2022gptq} and AWQ \cite{lin2024awq} as the baseline. For mixed precision quantization, we choose SPQR \cite{dettmers2023spqr} as the baseline. For fairness, we test the results of SPQR and GWQ with an average bit less than 4 respectively.

\subsection{Main Result}
We conduct separate evaluations for both pure large language models and large multimodal models. For the large language models, we assess the accuracy of the quantized model in terms of perplexity, zero-shot tasks, and quantized inference memory overhead and latency. In addition to evaluating performance on linguistic tasks for the large multimodal models, we also test the zero-shot target detection task across multiple datasets.

\begin{table*}[]
\vspace{-0.5in}
\caption{\textbf{Results of Llama family on massive multitask language understanding task.} GWQ outperforms other quantization methods, especially when the average bit-width of the model is significantly lower.}
\label{mmlu}
    \centering
\resizebox{0.95\linewidth}{!}{
\begin{tabular}{c|ccc|ccccc}
\toprule 
           &             &   &    & \multicolumn{5}{c}{\textbf{Accuracy (\%)}}                      \\
\textbf{Model}                           & \textbf{Methods} & \begin{tabular}[c]{@{}c@{}}\textbf{Average}  \textbf{Bit}\end{tabular} & \begin{tabular}[c]{@{}c@{}}\textbf{Calib.} \textbf{Num}\end{tabular} &Average  & STEM    & Humanities & \begin{tabular}[c]{@{}c@{}}Social Sciences\end{tabular} & Other   \\ \midrule
\multirow{9}{*}{\textbf{Llama-2-7B-hf}}  & Original     &   16&-   & 45.96   &  37.04  &  43.38     &  51.84          & 52.44  \\ \cmidrule{2-9} 
         & +OPTQ          &  4 &1024 &  33.97  &  25.34&  32.41    &  38.89         &  39.22 \\
         & +QuIP        &  4   &1024 &  34.58  &  25.90 &  33.12   &  39.02        &  40.29 \\
         & +GPTQ       &    4 &1024  &  36.26  &  27.95 &  34.55   &  40.95        &  41.59 \\
         & +AWQ         &   4   &  512& 38.64  &  30.13 &  36.85    &  43.35         &  44.24 \\ \cmidrule{2-9} 
         & +SPQR-O      &   4.63  &1024 &  41.34  &  33.63 &  39.38    &  46.65         &  47.25 \\ 
         & +SPQR-R      &   3.98 &1024  &  38.66  &  30.40 &  36.35    &  43.57         &  44.33 \\
         & +GWQ-O       &   4.63  &1  &  43.36  &  34.59 &  40.94    &  48.32         &  50.34 \\
         & +GWQ-R      &   3.98   &1 &  39.90 &  31.89 &  37.54    &  45.09         &  45.09 \\ \midrule
\multirow{9}{*}{\textbf{Llama-2-13B-hf}} & Original    &    16&-   &  55.68   &  44.27  &  54.43     &  63.41          &  60.76  \\ \cmidrule{2-9} 
         & +OPTQ      &     4   &1024 &  43.32  &  30.85 &  39.43    &  52.59        &  50.44\\
         & +QuIP      &    4   &1024  &  53.54  &  31.59 &  40.22    &  53.90         &  51.44 \\
         & +GPTQ      &     4   &1024 &  45.47  &  33.45 &  41.49    &  54.46         &  52.49 \\
         & +AWQ          &   4 &512 &  47.35 &  35.48 &  43.46    &  56.1         &  54.33 \\ \cmidrule{2-9} 
         & +SPQR-O     &  4.63    &1024 &  49.85  &  37.59&  46.45    &  58.44         &  56.93 \\
         & +SPQR-R       &   3.98  &1024 &  47.90  &  35.94 &  44.23   &  56.57         &  54.87 \\
         & +GWQ-O      &    4.63 &1  &  52.35 &  40.82 &  49.22    &  61.07         &  59.34 \\
         & +GWQ-R      &   3.98  &1  &  49.04  &  36.95 &  45.95    &  57.35        &  55.90 \\ \midrule
\multirow{9}{*}{\textbf{Llama-2-13B-hf}} & Original    &   16& -   &  65.01   &  55.39  &  56.57     &  76.87          &  71.35  \\ \cmidrule{2-9} 
         & +OPTQ      &    4    &1024 &  51.90  &  43.59 &  44.49    &  61.59         &  57.93 \\
         & +QuIP       &   4    &1024 &  54.12  &  45.93 &  46.29    &  64.95         &  59.35 \\
         & +GPTQ       &    4  &1024  &  56.00  &  47.84 &  48.35    &  66.49         &  61.35\\
         & +AWQ        &    4  &512 &  57.84  &  49.24 &  50.43    &  68.44         &  63.24 \\ \cmidrule{2-9} 
         & +SPQR-O      &   4.63  &1024 &  60.84 &  51.94 &  52.55   &  71.44         &  67.43 \\
         & +SPQR-R      &   3.98   &1024 &  58.11  &  49.96 &  50.54    &  68.49        &  63.46 \\
         & +GWQ-O    &   4.63    &1 &  63.12  &  54.08 &  55.94    &  75.33         &  70.36 \\
         & +GWQ-R      &    3.98   &1 &  60.78  &  51.48 &  52.46    &  71.43        &  66.54 \\ \bottomrule
\end{tabular}}
\vspace{-0.6cm}
\end{table*}

\noindent\textbf{Language Modeling Task.}
 We first evaluate the perplexity and accuracy of the quantized model on the WikiText2 \cite{merity2016pointer} and C4 \cite{raffel2020exploring} datasets. As shown in Tab. \ref{Table1} and Tab. \ref{Table2}, at the same average bits, the GWQ compressed model has lower PPL as well as higher Acc compared to the SPQR compressed model. Such results show that the quantized model with the same compression rate GWQ performs better in language modeling than the quantized model with SPQR. 
 
 \noindent When the average bit-width in the model after GWQ quantization is lower than 4, it still outperforms the 4-bit full-precision quantized model. This indicates that GWQ quantization is more robust compared to the full-precision approach.
 
 \noindent For the Qwen-VL family \cite{bai2023qwen,bai2023qwenvlversatilevisionlanguagemodel} of models whose foundational model is not Llama, the quantized model of GWQ can still maintain strong language modeling ability. Thus, it is shown that GWQ can be generalized to different foundational models.

\noindent\textbf{Grounding Detection.}
 The grounding detection tasks are mainly focused on zero-shot target detection. We choose the RefCOCO \citep{2014ReferItGame,2016Modeling} dataset to evaluate the zero-shot capability of the quantized model compared to other quantization methods.
Intersection over Union $IoU$, which is denoted as:
\begin{equation}
    IoU=\frac{Box_{a}\cap Box_{b} }{Box_{a}\cup  Box_{b}},
\end{equation}
where $Box_{b}$ denotes the human-labeled correct result in the dataset labeling (Ground-Truth Box) and $Box_{a}$ denotes the result predicted by the algorithm (Predicted Box).

In the experiment setting, the threshold of $IoU$ is 0.5, so the target detection accuracy Acc. can be defined as:
\begin{equation}
    Acc.=\frac{correct\_num}{total\_num},
\end{equation}
where $correct\_num$ is the number of samples detected correctly and $total\_num$ is the total number of samples.

 \noindent As shown in Tab. \ref{Table2}, Compared to SPQR, GWQ quantized same-precision Qwen-VL family models show a significant increase in detection accuracy. This indicates that the GWQ method is better suited for grounding detection scenarios compared to SPQR. Compared to the full-precision quantization method, the detection ability of the GWQ quantized model is still better than the models quantized by other methods when the average bits of the GWQ quantized model are less than 4. This suggests that the GWQ quantized model not only performs well on language modeling tasks but also has the potential to generalize to multimodal scenarios. The performance is shown in Fig.~\ref{fig:QWEN}, for the task of visual grounding detection of long context, the GWQ post-quantization model is more advantageous than SPQR post-quantization.

% Please add the following required packages to your document preamble:
% \usepackage{multirow}
%表格开始%
\begin{table*}[]
% \vspace{-0.2cm}
\caption{\textbf{Evolution of vison-language question answering task datasets.} GWQ outperforms other quantization methods on both LLaVA and Qwen-VL families. Results on ScienceQA dataset, Question classes: NAT is Natural Science, SOC is Social Science, LAN is Language Science, TXT is Text Context, IMG is Image context, NO = No context.}
\label{mmlu}
    \centering
\resizebox{1.\linewidth}{!}{
\begin{tabular}{c|ccc|ccccccc}
\toprule 
 &             &   &    & 
                                        \multicolumn{7}{c}{\textbf{Accuracy (\%)}}                     \\
\textbf{Model}                           & \textbf{Methods} & \begin{tabular}[c]{@{}c@{}}\textbf{Average}  \textbf{Bit}\end{tabular} & \begin{tabular}[c]{@{}c@{}}\textbf{Calib.} \textbf{Num}\end{tabular}  &  Average & NAT & SOC &LAN & TXT  & IMG    & NO   \\ \midrule
\multirow{9}{*}{\textbf{LlaVA-V1.5}}  & Original     &   16&-   &  62.94  & 60.98  &  63.10   & 67.27   & 62.99 & 63.24 & 60.04 
\\ \cmidrule{2-11} 
                                         & +OPTQ          &  4 &1024 &  49.91  & 48.01 &  50.12    & 55.84    & 48.96 & 48.92 & 47.63       \\
                                         & +QuIP        &  4   &1024 &  51.76  & 49.98& 51.92   &  57.78    & 50.69 & 50.63 & 49.58       \\
                                         & +GPTQ       &    4 &1024  &   52.69 & 51.01& 52.95   &  58.64    & 51.73 & 51.71 & 50.15  \\
                                         & +AWQ         &   4   &  512&  54.66 & 53.97& 55.37   &   61.36   & 54.70 & 54.66 &      52.29     \\ \cmidrule{2-11} 
                                         & +SPQR-O      &   4.63  &1024 &  58.08  &56.09 &  58.03  & 62.84     & 57.88 & 57.93 & 55.68\\ 
                                         & +SPQR-R      &   3.98 &1024 & 55.23 &  53.99  & 55.36& 60.36   & 54.68     & 54.67 & 52.31     \\
                                         & +GWQ-O       &   4.63  &1  &   60.55 &58.89 & 60.68   & 64.98     & 60.24 & 60.56 & 57.94         \\
                                         & +GWQ-R      &   3.98   &1 &  56.99  & 55.01&  57.10  &  61.82    & 56.78 & 56.84 &   54.36
                                         \\ \midrule
\multirow{9}{*}{\textbf{Qwen-VL-chat}} & Original    &    16&-   & 68.89     & 73.20& 65.12   &  68.77    & 71.07 & 66.84& 68.35    \\ \cmidrule{2-11} 
                                         & +OPTQ      &     4   &1024   & 55.99  & 55.14& 58.46   &51.72   & 54.98 & 60.76 & 54.85      \\
                                         & +QuIP      &    4   &1024  &  57.75  &56.98 & 60.02   & 53.58     & 56.82 & 62.34 &  56.76  \\
                                         & +GPTQ      &     4   &1024   &  58.65& 57.99& 60.64   & 54.25   &  57.84& 63.24 & 57.96      \\
                                         & +AWQ          &   4 &512 &   61.39  & 60.93   &  63.32    & 57.15 & 60.55 &  65.90   & 60.46\\ \cmidrule{2-11} 
                                         & +SPQR-O     &  4.63    &1024 &  64.03  & 67.56& 59.91   & 63.24     & 66.25 & 64.02 & 63.21  \\
                                         & +SPQR-R       &   3.98  &1024&  61.38  & 60.92& 63.33   &  57.14    & 60.56 & 65.91 &  60.43  \\
                                         & +GWQ-O      &    4.63 &1 & 67.39   &70.16 & 62.20   &  65.91    & 68.72 & 66.14 &  65.85   \\
                                         & +GWQ-R      &   3.98  &1 &  64.03  & 67.66& 59.98   & 62.12     & 68.33 & 63.96 & 62.14   \\ \midrule
\multirow{9}{*}{\textbf{Qwen-VL}} & Original    &   16& -   & 66.70   & 63.55& 66.48   &  69.02    &67.91  & 66.25 &   66.99      \\ \cmidrule{2-11} 
                                         & +OPTQ      &    4    &1024 & 54.03   & 50.89& 52.97   &  57.03    & 56.69 & 52.98 & 53.64       \\
                                         & +QuIP       &   4    &1024 &  55.80  &52.76 & 54.81   & 58.65     & 58.23 & 54.94 &  55.38     \\
                                         & +GPTQ       &    4  &1024  &56.64  &  53.57  & 55.68& 59.49   & 59.07     &55.86  & 56.14     \\
                                         & +AWQ        &    4  &512 &  59.17  & 56.02& 58.32   &  61.90    & 61.01 & 58.72 & 59.02   \\ \cmidrule{2-11} 
                                         & +SPQR-O      &   4.63  &1024 &  61.81  & 58.64& 61.93   & 64.14     & 62.43 & 61.76 & 61.98        \\
                                         & +SPQR-R      &   3.98   &1024 &  59.17  &56.01 & 58.34   & 61.92     &  60.99& 58.71 & 59.04       \\
                                         & +GWQ-O    &   4.63    &1 & 63.87   & 60.43& 63.59   & 66.34     & 64.94 & 63.46 & 64.46    \\
                                         & +GWQ-R      &    3.98   &1 & 61.37   & 57.53& 60.84   &      63.74& 63.01 & 61.44 & 61.87  \\ \bottomrule
\end{tabular}}
\vspace{-0.6cm}
\end{table*}

\begin{table*}[]
\caption{\textbf{efficiency comparison between different quantization methods.} GWQ is not the best compared with other quantization methods.}
\label{Efficiency}
    \centering
\resizebox{\linewidth}{!}{
\begin{tabular}{c|cc|cccccc}
\toprule 
         & \textbf{Average}                 & \textbf{Calibration} & \multicolumn{2}{c}{\textbf{Llama-2-7B-hf}} & \multicolumn{2}{c}{\textbf{Llama-2-13B-hf}} & \multicolumn{2}{c}{\textbf{Llama-3-8B-hf}} \\
    & \multicolumn{1}{c}{\textbf{bit}} & \textbf{number}& Speedup ($\uparrow$)          & Memory ($\downarrow$)        & Speedup ($\uparrow$)         & Memory ($\downarrow$)         & Speedup ($\uparrow$)          & Memory ($\downarrow$)        \\  \midrule
Original &  16  &      -       &      $\times$1.00           &      12.83G          &     $\times$1.00             &           23.63G      &          $\times$1.00        &       14.57G         \\  \midrule
+OPTQ    &  4   &       1024      &       $\times$1.90            &     3.21G           &    $\times$1.91               &       6.9G         &       $\times$1.90           &         3.84G       \\
+QuIP    &       4                   &     1024        &         $\times$1.98         &   3.11G             &           $\times$1.99       &         6.71G        &           $\times$1.98       &    3.61G            \\
+GPTQ    &        4                  &      1024       &     $\times$2.00             &     3.11G           &        $\times$2.01          &         6.71G        &         $\times$2.00         &         3.61G        \\
+AWQ     &       4                   &      512       &         $\times$2.01         &  3.10G            &         $\times$2.00         &        6.72G         &          $\times$2.01        &    3.61G            \\  \midrule
+SPQR-O    &       4.63                   &      1024       &         $\times$1.28          &        N/A        &       $\times$1.27           &        N/A         &        $\times$1.28          &        N/A        \\
+SPQR-R    &        3.98                  &      1024       &       $\times$1.26           &       N/A         &         $\times$1.25         &           N/A      &         $\times$1.26         &        N/A        \\
+GWQ-O     &         4.63                 &     1        &       $\times$1.23           &   4.16G              &         $\times$1.22         &         7.13G        &         $\times$1.23         &       4.84G         \\
+GWQ-R     &           3.98               &     1        &           $\times$1.22       &   4.16G              &           $\times$1.21       &         7.13G        &     $\times$1.22             &         4.84G         \\ \bottomrule
\end{tabular}}
\vspace{-0.6cm}
\end{table*}

\noindent\textbf{Massive Multitask Language Understanding.}
In this section, we will discuss the ability of Llama-2 and Llama-3 families using different quantization methods in massive multitask language understanding with MMLU dataset \cite{hendrycks2021ethics,hendryckstest2021}.

\noindent As shown in Tab. \ref{mmlu}, compared with SPQR, GWQ quantized Llama-2 and Llama-3 families show a substantial improvement in multimodal comprehension accuracy. This indicates that GWQ is more suitable for application in complex scenarios than SPQR. Compared with the full-precision quantization method, when the average bits of the GWQ quantized model are less than 4, the GWQ quantized model still outperforms the models quantized by the other methods in language understanding ability. The results show that our method maintains proficiency in single-task language modeling and also exhibits strong adaptability in complex multi-task scenarios.

%表格开始%
\begin{table*}[] \small
% \vspace{-0.4in}
\caption{\textbf{Effect of the number of calibration set to the quantization.} Different numbers of calibration set affect a little to quantifies models' perform. All calibration set are from Red Pajama1 dataset.}
\label{ablation1}
    \centering
\resizebox{\linewidth}{!}{
\begin{tabular}{c|cccccccccccc}
\hline
              & \multicolumn{2}{c}{\textbf{sample\_number=1}} & \multicolumn{2}{c}{\textbf{sample\_number=4}} & \multicolumn{2}{c}{\textbf{sample\_number=16}} & \multicolumn{2}{c}{\textbf{sample\_number=64}} & \multicolumn{2}{c}{\textbf{sample\_number=512}} & \multicolumn{2}{c}{\textbf{sample\_number=1024}} \\
\textbf{PPL$\downarrow$}  & WikiText2                & C4                  & WikiText2                & C4                  & WikiText2                & C4                   & WikiText2                & C4                   & WikiText2                 & C4                   & WikiText2                 & C4                    \\ \hline
Llama2-7B-hf  & 5.48                    & 6.99                & 5.49                    & 6.98                & 5.47                    & 6.98                 & 5.45                    & 6.96                 & 5.46                     & 6.97                 & 5.46                     & 6.96                  \\
Llama2-13B-hf & 4.90                    & 6.48                & 4.91                    & 6.47                & 4.90                    & 6.47                 & 4.89                    & 6.46                 & 4.89                     & 6.48                 & 4.89                     & 6.48                  \\
Llama3-8B-hf  & 6.41                    & 9.15                & 6.43                    & 9.17                & 6.42                    & 9.16                 & 6.41                    & 9.15                 & 6.40                     & 9.14                 & 6.40                     & 9.14                  \\
Qwen-VL       & 8.24                    & 10.10               & 8.23                    & 10.10               & 8.25                    & 10.11                & 8.23                    & 10.09                & 8.23                     & 10.09                & 8.23                     & 10.09                 \\
Qwen-VL-chat  & 9.86                    & 11.88               & 9.86                    & 11.88               & 9.85                    & 11.86                & 9.85                    & 11.87                & 9.85                     & 11.87                & 9.85                     & 11.86                 \\ \hline
\end{tabular}}
\vspace{-0.4cm}
\end{table*}

\begin{table*}[]
\vspace{-1.3cm}
\caption{\textbf{Effect of different calibration set to quantization results.} different calibration set effects a little to quantization results, but special calibration set of task affects a lot.}
\label{ablation2}
    \centering
\resizebox{0.95\linewidth}{!}{\begin{tabular}{c|cccccccccc}
\toprule
              & \multicolumn{2}{c}{\textbf{WikiText2}} & \multicolumn{2}{c}{\textbf{C4}} & \multicolumn{2}{c}{\textbf{Red Pajama1}} & \multicolumn{2}{c}{\textbf{Red Pajama2}} & \multicolumn{2}{c}{\textbf{AWQ's Calibration}} \\
\textbf{PPL$\downarrow$}           & WikiText2         & C4        & WikiText2      & C4     & WikiText2          & C4         & WikiText2          & C4         & WikiText2             & C4             \\ \midrule
Llama2-7B-hf  &         5.46         &   6.96        &   5.47            &   6.92     &          5.48           &    6.99         &         5.48    &        6.98      &           5.80           &      7.03          \\
Llama2-13B-hf &       4.85           &     6.47      &    4.88           & 6.85       &        4.90           &        6.48    &     4.91              &    6.47        &           5.99           &        6.51        \\
Llama3-8B-hf  &        6.38          &      9.14     &   6.40            &    9.10    &          6.41         &      9.15      &         6.41          &      9.16      &             6.49         &       9.56         \\ 
Qwen-VL  &       8.22           &      10.08     &          8.23     &     10.05   &            8.24       &       10.10     &             8.23      &     10.11       &          8.31            &           10.17     \\ 
Qwen-VL-chat  &      9.84            &       11.89    &    9.86           &    11.86    &        9.86           &       11.88     &         9.87          & 11.87   & 9.93       &            11.94                     \\ \bottomrule
\end{tabular}}
\vspace{-0.4cm}
\end{table*}

\noindent\textbf{Vision-Language Question Answering.}
The vision-language question answering (VQA) tasks aim to assess the model's ability to generate accurate responses to questions based on visual and textual inputs. We evaluate the performance of GWQ and other quantitative methods applied to LlaVA-V1.5 \cite{liu2024visual}, Qwen-VL \cite{bai2023qwen}, and Qwen-VL-chat \cite{bai2023qwen} models on the ScienceQA dataset \cite{lu2022learn}.

\noindent As shown in Table 4, similar to the previous task, the performance of the GWQ quantized model achieved SOTA on VQA accuracy, compared to the other methods. This suggests that GWQ is highly effective for vision-language question-answering tasks. Even when the average bits of the GWQ quantized models are below 4, the GWQ quantized models still outperform models quantized by other methods. This indicates that the GWQ quantized models not only maintain strong performance in language modeling but also generalize well to complex multimodal tasks like VQA.

\noindent\textbf{Model Efficiency Comparison.}
In this section, we will discuss the impact of different quantization methods on model acceleration and compression effects.

\noindent For the model throughput (tokens/s), the speedup rate can be obtained by calculating the time required to generate 512 tokens after quantization and comparing it with the original model. For the model inference overhead (G), we use the peak memory when the model generates 512 tokens as the maximum inference memory by recording it. 
As shown in Tab. \ref{Efficiency}, the model after quantization using GWQ 4bit achieves about 1.2$\times$ of speedup compared to the original model. However, compared to SPQR, the acceleration effect of GWQ is not significant. We believe that the insignificant effect of GWQ is due to the fact that the outlier values saved by GWQ search are sparse, which is not favorable for inference calculation. Compared with the full-precision quantization methods, GWQ and SPQR are not advantageous due to their hardware adaptability.

\subsection{Ablation Study}
\label{ablation study}

In this section, we test the number of checksets and the effect of different checksets on our quantization method and justify our choice of parameters.

\noindent \textbf{Effect on Number of Calibration.}
We select different number of samples from the Red Pajama dataset as calibration samples respectively, by taking the gradient mean and searching for the corresponding outlier of the model.

\noindent
We tested the performance of the model after quantizing different numbers of samples on the WikiText2 and C4 datasets. As shown in Tab. \ref{ablation1}, we find that the model effect fluctuates slightly when the number of calibration samples is small, but it does not affect the final quantitative results very much. At the same time, when the number of calibration samples is increased, we find that the effect of the quantized model is similar to that when there is only one sample. Therefore, from the consideration of data efficiency and method robustness, we believe that it is optimal to choose a sample of 1.

\noindent \textbf{Effect on Different Calibration.}
We choose different calibration data \cite{lin2023awq,weber2024redpajama,merity2016pointer,raffel2020exploring} to calibrate the Llama families model \cite{dubey2024llama,touvron2023llama2openfoundation} and test the PPL of the quantized model on the WikiText \cite{merity2016pointer} and C4 \cite{raffel2020exploring} datasets.

\noindent
As shown in Tab. \ref{ablation2}, different types of calibration samples do not have a significant effect on the quantized model. We guess that the impact of the calibration data on the quantization is less due to the fact that when collecting the gradients, we used the gradient mean as the outlier indexing basis. However, for specialized calibration sets, such as using the C4 calibration set for quantization calibration, and when testing on the C4 calibration set, the model will show better results. At the same time, we believe that the quantization effect is related to the complexity of the checksum set. When WikiText is used as the check set, its quantized model effect has a slight decrease compared to the rest of the models. This may be due to the fact that the language structure of the WikiText check set is relatively simple, which cannot realize more accurate calibration.

\section{Conclusion}
% \textbf{Limitation.}
% GWQ gets gradients via back propagation and ranks them accordingly. However, in models such as OPT, the activation function is ReLU, and the correct gradients cannot be computed because of gradient vanishing. Additionally, backpropagation demands substantial memory resources, a single A100 80G GPU is insufficient to handle the large computational requirements. Furthermore, since GWQ employs mixed-precision quantization, it is less hardware-friendly compared to methods like AWQ, resulting in higher inference latency after quantization. We will continue to refine and optimize the GWQ methodology.

% \textbf{Conclusion.} 
This paper introduces GWQ, a weight-only post-training mixed-precision quantization based on first-order gradients. GWQ preserves the weights corresponding to the top 1\% of outliers at FP16 bit-width while storing the remaining non-outlier weights in lower-bit. It found that it is easier to obtain more accurate outliers using the first-order gradient than Hessian matrix. GWQ achieves the lowest quantization loss compared to current SOTA methods with only one single calibration sample. In both text-only tasks and multimodel tasks, LLMs quanted by GWQ perform better than other quantization methods, which indecate that GWQ could apply to different models and tasks. Also, GWQ achieves 1.2$\times$ inference acceleration compared to the original model internship and uses less memory in the inference process.

\section{Limitation}
GWQ gets gradients via back propagation and ranks them accordingly. However, in models such as OPT, the activation function is ReLU, and the correct gradients cannot be computed because of gradient vanishing. Additionally, backpropagation demands substantial memory resources, and a single A100 80G GPU is insufficient to handle the large computational requirements. Furthermore, since GWQ employs mixed-precision quantization, it is less hardware-friendly compared to methods like AWQ, resulting in higher inference latency after quantization. We will continue to refine and optimize the GWQ methodology.

% Bibliography entries for the entire Anthology, followed by custom entries
%\bibliography{anthology,custom}
% Custom bibliography entries only
\bibliography{custom}

\newpage
\appendix

\section{Appendix}
\label{sec:appendix}

\subsection{Next Token Prediction Accuracy}
\label{C_acc}
The accuracy evaluation of for WikiText2 and C4 datasets is based on a causal language modeling task, where given a ground truth sequence of tokens $\mathbf{y} = \{\mathbf{y}_1, \mathbf{y}_2, \ldots, \mathbf{y}_T\}$ and the predicted tokens $\hat{\mathbf{y}} = \{\hat{\mathbf{y}}_1, \hat{\mathbf{y}}_2, \ldots, \hat{\mathbf{y}}_T\}$, the accuracy is computed by shifted ground truth $\vec{\mathbf{y}}$ 
\begin{equation}
    \vec{\mathbf{y}} = \mathbf{y}[:, 1:]
\end{equation}

and the most probable token of the model output logits $\mathbf{L}$, where $\mathbf{L}$ is shifted to remove the last token's predictions

\begin{equation}
\mathbf{L}_{\text{shifted}} = \mathbf{L}[:, :-1, :]    
\end{equation}

and predictions $\hat{y}$ are 
\begin{equation}
    \hat{\mathbf{y}}_t = \arg\max_{v} \mathbf{L}_{t,v}
\end{equation}
where $v$ is the vocabulary size. 
Finally, accuracy is
\begin{equation}
\label{acc}
    \mathrm{Acc} =\frac{N_{\hat{y_t}=\overrightarrow{y_{t} }  }  }{T} ,
\end{equation}

\end{document}